\documentclass[conference]{IEEEtran}
\IEEEoverridecommandlockouts

\usepackage{cite}
\usepackage{amsmath,amssymb,amsfonts}
\usepackage{algorithmic}
\usepackage{graphicx}
\usepackage{textcomp}
\usepackage{xcolor}
\usepackage{stfloats}
\usepackage{booktabs}
\usepackage{multirow}
\usepackage{makecell}
\usepackage{caption}
\usepackage{float}
\usepackage{placeins}
\usepackage{hyperref}
\usepackage[letterpaper,margin=0.75in]{geometry}
\def\BibTeX{{\rm B\kern-.05em{\sc i\kern-.025em b}\kern-.08em
    T\kern-.1667em\lower.7ex\hbox{E}\kern-.125emX}}

\newcommand{\pitl}{planner-in-the-loop}
\newcommand{\Atoms}{\mathrm{Atoms}}

\begin{document}

\title{Toward Secure and Reliable PDDL Formalization of Large Language Models with Planner-in-the-Loop Feedback
}

%\author{\IEEEauthorblockN{Anonymous Authors}}

 \author{
 \IEEEauthorblockN{Jiamei Jiang\textsuperscript{1,2}, Jiajing Zhang\textsuperscript{1,2}, Feifei Mo\textsuperscript{1,3}, Linjing Li\textsuperscript{1*}, Daniel Zeng\textsuperscript{1}}
\IEEEauthorblockA{\textsuperscript{1}State Key Laboratory of Multimodal Artificial Intelligence Systems, \\ Institute of Automation, Chinese Academy of Sciences, Beijing, China}
 \IEEEauthorblockA{\textsuperscript{2}School of Artificial Intelligence, University of Chinese Academy of Sciences, Beijing, China}
 \IEEEauthorblockA{\textsuperscript{3}School of Industry-education Integration, University of Chinese Academy of Sciences, Beijing, China}
\IEEEauthorblockA{jiangjiamei2024@ia.ac.cn, jiajing.zhang@ia.ac.cn, mofeifei2025@ia.ac.cn, linjing.li@ia.ac.cn, dajun.zeng@ia.ac.cn}
}

\maketitle
\begin{abstract}
Planning often requires symbolic specifications that are both executable and verifiable. For large language models deployed in autonomous or decision-support systems, failures in such formalization may lead to unverifiable decisions, execution failures, or unsafe downstream behavior. We present NL-PDDL-Bench, a multi-domain benchmark for natural-language-to-PDDL specification construction with planner-verified executability and controlled difficulty scaling by object count. We further propose a planner-in-the-loop framework that uses validator and planner diagnostics to revise non-executable specifications through localized edits. Building on this infrastructure, we develop a planner-grounded optimization recipe that combines parameter-efficient Low-Rank Adaptation supervised fine-tuning, offline planner-derived preference pairs for Direct Preference Optimization, and inference-time planner-in-the-loop repair, without requiring online planner calls during training. We also provide a unified evaluation suite for parseability, solvability, specification similarity, and outcome-aware plan-level consistency against planner references. Experiments on representative model families show substantial gains in planner success and plan-level agreement, with improved robustness under difficulty scaling and cross-domain variation. These results highlight the value of externally verifiable formalization for reliable deployment of LLMs in safety- or security-sensitive planning systems. Code and data are available at: \url{https://github.com/ibasicplan/NL-PDDL-Bench}
\end{abstract}


% =========================
% =========================
\section{Introduction}

Planning is a capability in artificial intelligence. Given an initial state and a set of goal conditions, a planner can synthesize an executable sequence of actions for applications such as robot control, autonomous driving, and logistics scheduling. Planning is difficult because it demands faithful action models and satisfaction of reachability constraints over long horizons and large action spaces \cite{bercher2025survey,pozo2025abstraction}.

As large language models are used as interfaces for autonomous, embodied, and decision-support systems, formalization reliability becomes important for safe deployment \cite{yao2024llm_security_privacy}. Semantically inconsistent specifications can propagate into unverifiable decisions, execution failures, or unsafe downstream behavior. This motivates a planner-grounded perspective in which LLM outputs are externally checkable, executable, and corrigible under formal constraints.

Classical symbolic planners address these requirements through explicit, checkable specifications. PDDL is a widely adopted formalism that represents planning domains and problem instances in a structured, planner-verifiable form \cite{aeronautiques1998pddl}. When specifications are correct, state transitions and goal satisfaction can be validated by established planners and verification tools, yielding reproducible executability guarantees \cite{helmert2006fast,howey2004val}. However, constructing such specifications is costly and typically requires expert knowledge, which limits scalability and practical deployment.

Recent large language models (LLMs) offer strong language understanding and structured generation abilities, motivating their use to reduce modeling effort \cite{valmeekam2023planning,tantakoun2025formalizers,li2025unlocking}. Existing approaches largely follow two paradigms: direct action-sequence generation from task descriptions, and symbolic specification generation followed by symbolic planning \cite{planning_in_the_dark,env_interaction,gestrin2024nl2plan}. Although both can produce plausible outputs, reliability under planner verification remains the key obstacle.

For direct plan generation, actions that look reasonable often violate preconditions, break state consistency, or miss necessary constraints, and therefore fail under established planners \cite{planbench,acpbench}. More broadly, current LLMs remain far behind strong symbolic planners in correctness and reliability on classical planning tasks, especially under strict reachability and long-horizon constraints. For symbolic specification generation, outputs may be syntactically well formed yet still unsolvable because of subtle modeling errors such as missing objects, incorrect predicate usage, or incomplete initial facts \cite{generalized_planning_llm}. These failure modes show that planning requires consistency, executability, and verifiability that cannot be ensured by surface plausibility alone.

We address this challenge by releasing NL-PDDL-Bench and a unified planner-grounded pipeline for executable specification construction, optimization, and evaluation at scale. It couples large-scale planner-verified NL--PDDL pairs with planner-in-the-loop feedback and planner-grounded training signals, enabling reproducible analysis and robust performance comparisons across domains, difficulty levels, and variation settings.

\noindent\textbf{Contributions:}
\begin{itemize}
\item We propose a planner-in-the-loop feedback framework that turns validator and planner diagnostics into localized guidance for minimal edits toward executability.
\item We release NL-PDDL-Bench, a multi-domain benchmark with planner-verified executable instances and controlled difficulty scaling by object count.
\item We introduce a planner-grounded optimization recipe for executable specification construction, combining supervised fine-tuning, planner-derived preference optimization, and inference-time repair.
\item We develop a unified planner-grounded evaluation protocol and demonstrate strong gains in solvability and plan-level consistency, supporting the reliable and externally verifiable deployment of LLMs in safety-sensitive planning systems.
\end{itemize}

% =========================
\section{Definitions and Background}
% =========================

\subsection{Planning Formulation}
We focus on deterministic classical planning. A planning domain is denoted by $D=(V,A)$, where $V$ is a finite set of predicate symbols, and $A$ is a set of action schemas with preconditions and effects. Given $D$, a problem instance specifies an object set $O$, an initial state $s_0$ as a set of ground atoms, and a goal condition $g_{\mathrm{true}}$ \cite{volkema1983problem}:
\begin{equation}
P=(D,O,s_0,g_{\mathrm{true}})
\end{equation}
A classical planner searches for a finite plan $\pi=(a_0,\ldots,a_n)$ such that each $a_t$ is applicable in $s_t$ and induces a valid transition
\begin{equation}
s_{t+1}=\gamma(s_t,a_t)
\end{equation}
with the terminal state satisfying $s_n\models g_{\mathrm{true}}$. Under the PDDL convention, the domain file declares predicates, types, and action schemas, while the problem file instantiates $O$, $s_0$, and $g_{\mathrm{true}}$.

\subsection{Introduction to PDDL}
PDDL is a widely used formalism for classical planning \cite{gerevini2020introduction}. A specification is separated into a domain file and a problem file: the domain file defines predicate symbols, object types, and action schemas with explicit preconditions and effects, while the problem file instantiates a concrete task by declaring objects, initial facts, and goal conditions. This decomposition yields a normalized state-action representation that can be parsed and solved by symbolic planners, enabling reproducible verification of executability and solution quality.

In this work, PDDL serves as the formal interface for connecting natural-language task descriptions with symbolic planning toolchains. A large language model is required to produce problem specifications that respect the predicate signatures and action constraints defined by the domain. Deviations such as missing object declarations, inconsistent predicate usage, or incomplete initial facts can render an instance invalid or unsolvable, and these failures are reliably exposed by planner-based validation and solving, providing grounded signals for diagnosis and correction.

% =========================
\section{Method}
% =========================

\begin{figure*}[t]
  \centering
  \includegraphics[width=\textwidth]{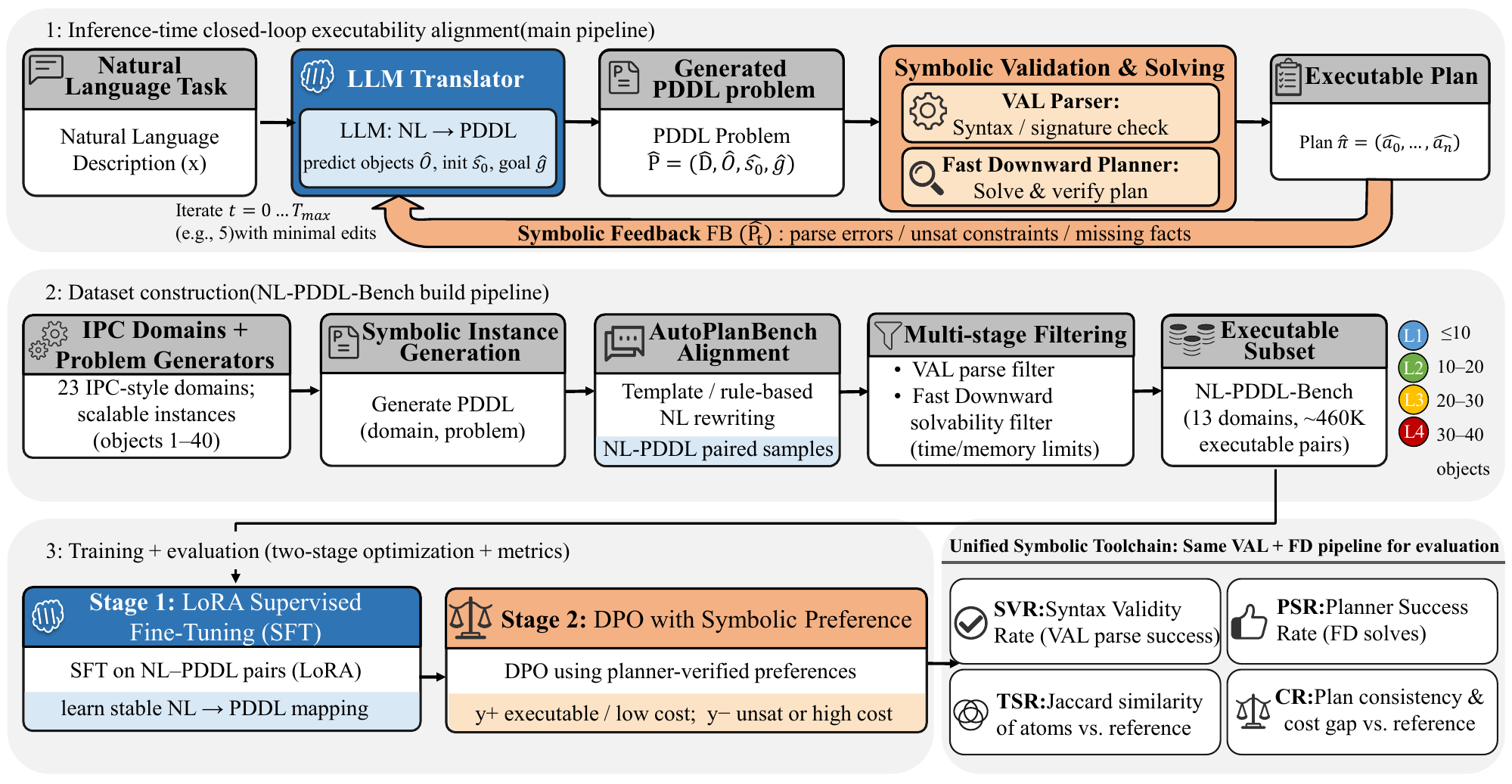}
  \caption{Overview of the planner-in-the-loop feedback framework.}
  \label{fig:framework}
\end{figure*}

Our method adopts a planner-in-the-loop framework that couples large language models with a standard symbolic planning toolchain to generate specifications that are parseable, solvable, and verifiable. The core idea is to place the planner in the generation loop: the model translates natural-language task descriptions into symbolic specifications, while the validator and planner provide reproducible evidence about well-formedness and solvability. Their diagnostics are then converted into targeted revision signals, enabling localized repairs under fixed symbolic semantics rather than relying on surface plausibility alone. To support systematic optimization and evaluation, we construct NL-PDDL-Bench and use a unified symbolic toolchain throughout dataset construction, verification, training, and testing. Each instance contains an aligned natural-language description, its corresponding specification, and planner-derived reference signals for measuring validity, solvability, and plan-level alignment. As illustrated in Fig.~\ref{fig:framework}, the overall workflow integrates benchmark construction, planner-grounded optimization, and diagnosis-guided refinement. On the learning side, we combine supervised fine-tuning with planner-grounded preference optimization; at inference time, if parsing or planning fails, the system converts diagnostic logs into structured feedback and prompts the model to perform localized repairs until verification succeeds or a fixed budget is reached.

\subsection{NL-PDDL-Bench}

NL-PDDL-Bench is designed to provide a reproducible and planner-verifiable foundation for training and evaluating natural-language specification construction. Rather than relying on offline text-level alignment or manual rewriting, we build an end-to-end automated pipeline that combines top-down task design with bottom-up symbolic verification, following the general benchmark-construction spirit of automatically deriving LLM planning evaluations from PDDL resources \cite{stein2023autoplanbench}. This pipeline enforces executable alignment: only instances that satisfy symbolic constraints and pass planner verification are included in the executable subset used for learning and evaluation, while failed instances yield structured diagnostic evidence under a unified toolchain. To make verification reproducible and diagnostically useful, we rely on planner-backed validation as a systematic check of semantic executability and inconsistency exposure \cite{goldman2012using}, and we adopt Fast Downward as the underlying solver backbone.

\begin{table}[htbp]
\centering
\small
\setlength{\tabcolsep}{4pt}
\renewcommand{\arraystretch}{1.15}
\caption{NL-PDDL-Bench statistics across difficulty levels.}
\label{tab:dataset_distribution}
\begin{tabular}{lccc}
\toprule
Level & Objects & Solvable pairs & \makecell{Plan length\\(median [Q1, Q3])} \\
\midrule
L1 & 1--10  & 154219 & 15.0 [12.8, 17.5] \\
L2 & 11--20 & 112003 & 39.0 [21.5, 91.5] \\
L3 & 21--30 & 93909  & 56.0 [19.0, 144.0] \\
L4 & 31--40 & 96918  & 86.5 [30.2, 208.5] \\
\bottomrule
\end{tabular}
\end{table}

We begin from 23 IPC-style source domains and their problem generators at the collection stage. For each symbolic instance, we generate an aligned natural-language description using deterministic structure-preserving templates that verbalize object declarations, initial facts, goal conditions, and domain-specific constraints directly from the corresponding PDDL problem file; no free-form paraphrasing or manual rewriting is involved. Difficulty is controlled by the total number of instantiated objects, and problems are stratified into four levels (L1--L4) covering 1--40 objects. After filtering domains with unstable generation or insufficient planner-verified coverage, the final released benchmark retains 13 domains: assembly, data-network, elevators, logistics, mprime, mystery, openstacks, rovers, snake, spider, tpp, woodworking, and zenotravel.

To guarantee executability and support planner-in-the-loop, we apply multi-stage verification. We first run a validator to filter specifications that violate syntax, predicate signatures, typing, or arity constraints. We then invoke Fast Downward under a fixed resource budget and retain only planner-solvable instances as the executable subset, recording planner-derived reference signals including plans and plan cost or length, as well as failure logs for diagnosis. This pipeline starts from approximately 1.707 million aligned NL--PDDL pairs and yields about 460 thousand planner-solvable high-quality instances. Table~\ref{tab:dataset_distribution} summarizes the resulting dataset distribution across difficulty levels and reports basic planning statistics.

\subsection{SFT and Planner-Grounded Preference Optimization}
We adopt a two-stage optimization scheme to improve executability and plan-quality alignment of generated specifications. The first stage performs parameter-efficient supervised fine-tuning to learn a stable mapping from text to problem specifications, emphasizing object declarations, type consistency, predicate signatures, and faithful encoding of initial and goal facts. This design is consistent with recent findings that targeted fine-tuning or instruction tuning can substantially strengthen LLM planning performance in structured symbolic settings \cite{li2025unlocking,verma2025pddlinstruct}. The second stage introduces planner-grounded preference optimization: we construct reusable preference signals via offline validation and planning, pushing the model toward specifications that are solvable and quality-aligned with reference solutions rather than merely well formed.

For SFT, we use LoRA \cite{hu2022lora} and learn a low-rank update on top of frozen base weights,
\begin{equation}
W = W_0 + \Delta W = W_0 + \alpha B_{\mathrm{LoRA}}A_{\mathrm{LoRA}}
\end{equation}
and minimize the standard negative log-likelihood,
\begin{equation}
\mathcal{L}_{\mathrm{NLL}} = -\frac{1}{T}\sum_{t=1}^{T}\log P_{\theta}(y_t \mid y_{<t}, x)
\end{equation}
Following common practice under constrained compute, we adopt QLoRA \cite{dettmers2023qlora}, which fine-tunes LoRA adapters on a frozen 4-bit quantized base model to reduce memory footprint while maintaining performance.

Supervision alone does not guarantee planner solvability. We therefore build an offline preference dataset: for each input $x$, the reference specification is treated as a positive sample $y^{+}$; candidate negatives are generated by programmatic perturbations and filtered by a validator and a planner to obtain $y^{-}$, which are either unsolvable or solvable but substantially worse than the reference in plan quality. We then apply Direct Preference Optimization \cite{rafailov2023direct} by optimizing the policy $f_{\theta}$ against a frozen reference $f_{\theta_0}$:
\begin{equation}
\begin{aligned}
\mathcal{L}_{\mathrm{DPO}}
&= \mathbb{E}\!\left[-\log \sigma(\beta \Delta)\right], \\
\Delta
&= \Bigl(\log f_{\theta}(y^{+}\!\mid x)-\log f_{\theta}(y^{-}\!\mid x)\Bigr) \\
&\quad - \Bigl(\log f_{\theta_0}(y^{+}\!\mid x)-\log f_{\theta_0}(y^{-}\!\mid x)\Bigr)
\end{aligned}
\end{equation}

Since preference labels are produced entirely by offline planner verification, this stage avoids planner calls during backpropagation while injecting stable symbolic signals for executability and quality alignment.

\subsection{Planner-in-the-Loop Feedback and Evaluation}
We build a verifiable neuro-symbolic loop by using a symbolic planner as a unified interface for parsing, solving, and diagnosis, and by placing it in the generation process at inference time. The model first produces a candidate problem specification; a validator and a planner then attempt to parse and solve it, returning structured diagnostics. When parsing or planning fails, we convert diagnostics into localized constraints and revision directives, and prompt the model to apply minimal edits while preserving the intended task semantics. The loop terminates once the specification becomes executable or the budget is exhausted. This design turns non-executable outputs into attributable symbolic defects and provides formally grounded evidence for both evaluation and optimization.

At test time, we create an isolated workspace per instance and record the domain file, the reference specification, intermediate model outputs, and planner logs. We first run a one-shot generation and invoke validation and planning. If parsing fails, we trigger localized repairs guided by validator errors; if parsing succeeds but planning fails, we provide feedback targeting structural omissions and consistency conflicts and continue iterative repairs. The final prediction is the best verified specification produced within the evaluation budget.

A typical failure case is a linguistically plausible but planner-invalid specification that omits a reachability-supporting fact. For example, in a Logistics instance, a generated problem may correctly declare the package, truck, initial location, and destination, yet omit a required connectivity fact between two locations. Such an output remains parseable but becomes unsolvable because no move action can be applied. In this case, planner feedback localizes the missing enabling condition, and \pitl\ revises the specification with a minimal edit that restores solvability.

Let $\mathcal{D}_{\mathrm{test}}=\{(x_i,y_i)\}_{i=1}^{N}$, where $x_i$ is the input description, $y_i$ is the reference specification, and $\hat{y}_i$ is the model output. We report four metrics that emphasize executable alignment.
Syntax Validity Rate measures whether $\hat{y}_i$ is parseable:
\begin{equation}
\mathrm{SVR}=\frac{1}{N}\sum_{i=1}^{N}\mathbf{1}\!\left[\mathrm{Parse}(\hat{y}_i)=\mathrm{OK}\right]
\end{equation}

Planner Success Rate measures solvability conditional on parseability. Define $I_{\mathrm{parse}}=\{i\mid \mathrm{Parse}(\hat{y}_i)=\mathrm{OK}\}$ and $I_{\mathrm{solve}}=\{i\mid \mathrm{Plan}(\hat{y}_i)=\mathrm{SUCCESS}\}$, then
\begin{equation}
\mathrm{PSR}=\frac{|I_{\mathrm{parse}}\cap I_{\mathrm{solve}}|}{|I_{\mathrm{parse}}|}
\end{equation}

Task Similarity Rate measures specification-level overlap in atomic facts using Jaccard similarity:
\begin{equation}
\mathrm{TSR}=\frac{1}{N}\sum_{i=1}^{N}\frac{|\Atoms(\hat{y}_i)\cap \Atoms(y_i)|}{|\Atoms(\hat{y}_i)\cup \Atoms(y_i)|}
\end{equation}
where $\Atoms(y)$ extracts the set of atomic facts from a specification $y$.

For plan-level alignment, we solve both the generated and reference problems under a planner configuration and obtain $(\mathrm{status}_i^{\mathrm{gen}},\pi_i^{\mathrm{gen}},c_i^{\mathrm{gen}})$ and $(\mathrm{status}_i^{\mathrm{ref}},\pi_i^{\mathrm{ref}},c_i^{\mathrm{ref}})$, where $c$ denotes plan cost or length. Consistency Rate is outcome-aware:
\begin{equation}
\begin{aligned}
\mathrm{CR}
&= \frac{1}{N}\sum_{i=1}^{N}
\mathbf{1}\!\Bigl[
\mathrm{status}_i^{\mathrm{gen}}
=
\mathrm{status}_i^{\mathrm{ref}}
\land
\Psi_i
\Bigr],
\\
\Psi_i
&=
\Bigl(
\mathrm{status}_i^{\mathrm{ref}}=\textsc{failure}
\Bigr)
\\
&\quad \lor
\Bigl(
|c_i^{\mathrm{gen}}-c_i^{\mathrm{ref}}|
\le \epsilon_i
\land
\pi_i^{\mathrm{gen}}\approx \pi_i^{\mathrm{ref}}
\Bigr)
\end{aligned}
\end{equation}

In our experiments, plan similarity is implemented using a commutativity-robust criterion based on normalized sequence edit similarity and bag-of-actions similarity, and cost tolerance is set relative to reference plan length.

% =========================
\section{Experiments}
% =========================
\subsection{Experimental Setup}
Experiments are conducted on NL-PDDL-Bench to evaluate executability alignment across multiple model families, thirteen planning domains, and stratified difficulty levels. All results are obtained under a unified symbolic toolchain in which a validator checks well formedness and Fast Downward evaluates solvability and returns reproducible plans and diagnostics under a fixed configuration: lazy greedy search with FF and landmark-sum heuristics under unit-cost adaptation, a 30-minute time limit, and a 30\,GB memory limit per instance \cite{helmert2006fast}. The test set is drawn from the executable subset of NL-PDDL-Bench, from which we sample one thousand planner-verified instances evenly across domains and difficulty strata, disjoint from training data; for planner-in-the-loop evaluation, the iteration budget is fixed to $T_{\max}=5$. We report executability metrics and plan-level alignment against planner-derived references, and analyze three aspects: comparison with direct action-sequence generation, attribution across training and repair settings, and robustness under difficulty scaling and cross-domain variation.

\subsection{Comparative Results}

\subsubsection{LLM-only Planning vs. Planner-Backed Formalization}
We compare end-to-end performance of direct LLM planning against our planner-backed formalization pipeline. This is not a like-for-like formalization-only comparison: our method first generates a PDDL specification and then delegates search to Fast Downward, whereas the proprietary baselines directly generate action sequences without an external symbolic planner. Our method uses a LoRA Llama 3.1-8B model with symbolic feedback to produce a planning specification, followed by Fast Downward to synthesize an executable plan. We report planner success rate on three domains. Logistics and Zenotravel use 100 instances each sampled from NL-PDDL-Bench for lightweight in-domain comparison, while Blocksworld uses 600 PlanBench-Mystery instances as an external stress test; PlanBench data are not used in NL-PDDL-Bench training.

\begin{table}[htbp]
\centering
\small
\setlength{\tabcolsep}{4pt}
\renewcommand{\arraystretch}{1.15}
\caption{Planner success rates across planning domains.}
\label{tab:direct_plan_generation}
\begin{tabular}{l c c c}
\toprule
Model & Logistics & Zenotravel & Blocksworld \\
\midrule
GPT-4o        & 22.4\% & 33.7\% & 0.0\%  \\
Claude 3.5    & 22.6\% & 57.9\% & 0.0\%  \\
GPT-5.2       & 42.3\% & 64.2\% & 6.2\%  \\
Claude 4.6    & 56.1\% & 97.9\% & 60.5\% \\
DeepSeek-R1   & 56.8\% & 98.9\% & 43.3\% \\
GLM-5         & 23.2\% & 75.8\% & 82.0\% \\
\textbf{Ours} & \textbf{78.0\%} & \textbf{99.0\%} & \textbf{87.3\%} \\
\bottomrule
\end{tabular}
\end{table}

Table~\ref{tab:direct_plan_generation} should therefore be read as a comparison between LLM-only planning and planner-backed formalization, rather than as an isolated measure of formalization quality alone. Direct generation can be competitive in individual domains, especially in Zenotravel, but remains substantially less stable across domains, with performance degrading sharply when tasks require stricter structural consistency and deeper state tracking, as in Blocksworld. By contrast, our approach uses a verifiable symbolic interface that delegates search to a planner and exploits planner-in-the-loop diagnostics for localized repair, attaining the best results on all three domains and remaining consistently strong across them.

\subsubsection{Main Results and Ablation Analysis}

We next unify overall comparison and ablation analysis under the same evaluation protocol. For each model, we report three settings: Baseline, where the base model directly generates a PDDL problem; SFT, which applies supervised fine-tuning without iterative symbolic repair at test time; and PITL, which augments the trained model with inference-time planner-in-the-loop feedback.

\begin{table}[htbp]
\centering
\small
\setlength{\tabcolsep}{4.2pt}
\renewcommand{\arraystretch}{1.13}
\caption{Main results and ablation on NL-PDDL-Bench.}
\label{tab:main_results}
\begin{tabular}{l l c c c c}
\toprule
Model & Setting & SVR & PSR & TSR & CR \\
\midrule
\multirow{3}{*}{Llama3.1-8B}
& Baseline & 76.3\% & 9.7\% & 52.9\% & 10.9\% \\
& SFT      & 85.7\% & 80.1\% & 82.5\% & 80.1\% \\
& PITL     & 84.9\% & 80.7\% & 82.8\% & 81.2\% \\
\addlinespace

\multirow{3}{*}{Qwen3-8B}
& Baseline & 82.1\% & 24.8\% & 61.5\% & 25.6\% \\
& SFT      & 82.1\% & 77.2\% & 81.2\% & 77.0\% \\
& PITL     & 82.7\% & 78.9\% & 80.8\% & 79.7\% \\
\addlinespace

\multirow{3}{*}{DeepSeek-R1-7B}
& Baseline & 84.6\% & 32.4\% & 34.1\% & 33.9\% \\
& SFT      & 88.7\% & 54.6\% & 55.2\% & 56.0\% \\
& PITL     & 90.2\% & 58.3\% & 57.0\% & 58.1\% \\
\addlinespace

\multirow{3}{*}{GPT-OSS-20B}
& Baseline & 69.9\% & 37.5\% & 39.9\% & 38.3\% \\
& SFT      & 66.9\% & 29.7\% & 29.2\% & 30.5\% \\
& PITL     & 79.0\% & 37.6\% & 31.5\% & 38.4\% \\
\bottomrule
\end{tabular}
\end{table}

Table~\ref{tab:main_results} reveals a consistent trend across the representative models: the principal gains are concentrated on PSR and CR rather than on SVR. This indicates that the main challenge is not merely syntactic well-formedness, but executable correctness under planner semantics. For Llama3.1-8B and Qwen3-8B, SFT already yields large improvements in solvability and plan-level agreement, showing that supervised learning effectively captures the core NL-to-PDDL mapping. PITL then provides additional gains, although smaller, by repairing residual local defects that remain invisible to text-only supervision.

DeepSeek-R1-7B follows the same pattern, with improvements again concentrated on planner-grounded metrics. This supports our broader claim that the framework improves executable alignment primarily by converting parseable-but-unsuccessful formalizations into solvable ones. The gain from PITL is therefore best interpreted as a symbolic correction effect rather than a purely linguistic one. This observation is also consistent with recent evidence that strong in-domain planning performance after fine-tuning does not necessarily translate into robust cross-domain generalization \cite{belcamino2026generalization_gap}.

GPT-OSS-20B behaves differently. Here, SFT degrades all four metrics relative to the baseline, indicating model-family sensitivity to the chosen parameter-efficient optimization recipe. Nevertheless, PITL recovers much of this loss and slightly improves PSR and CR over the baseline. This exception is informative: it suggests that PITL is robust as a correction mechanism, but its benefit is bounded by the quality of the underlying formalization model. Overall, the unified results show that SFT improves global formalization ability, while PITL serves as a module that strengthens executability through planner-grounded local repair.

\subsection{Analytical Experiments}

\subsubsection{Performance under Difficulty Scaling}

\begin{figure}[htbp]
  \centering
  \includegraphics[width=\columnwidth]{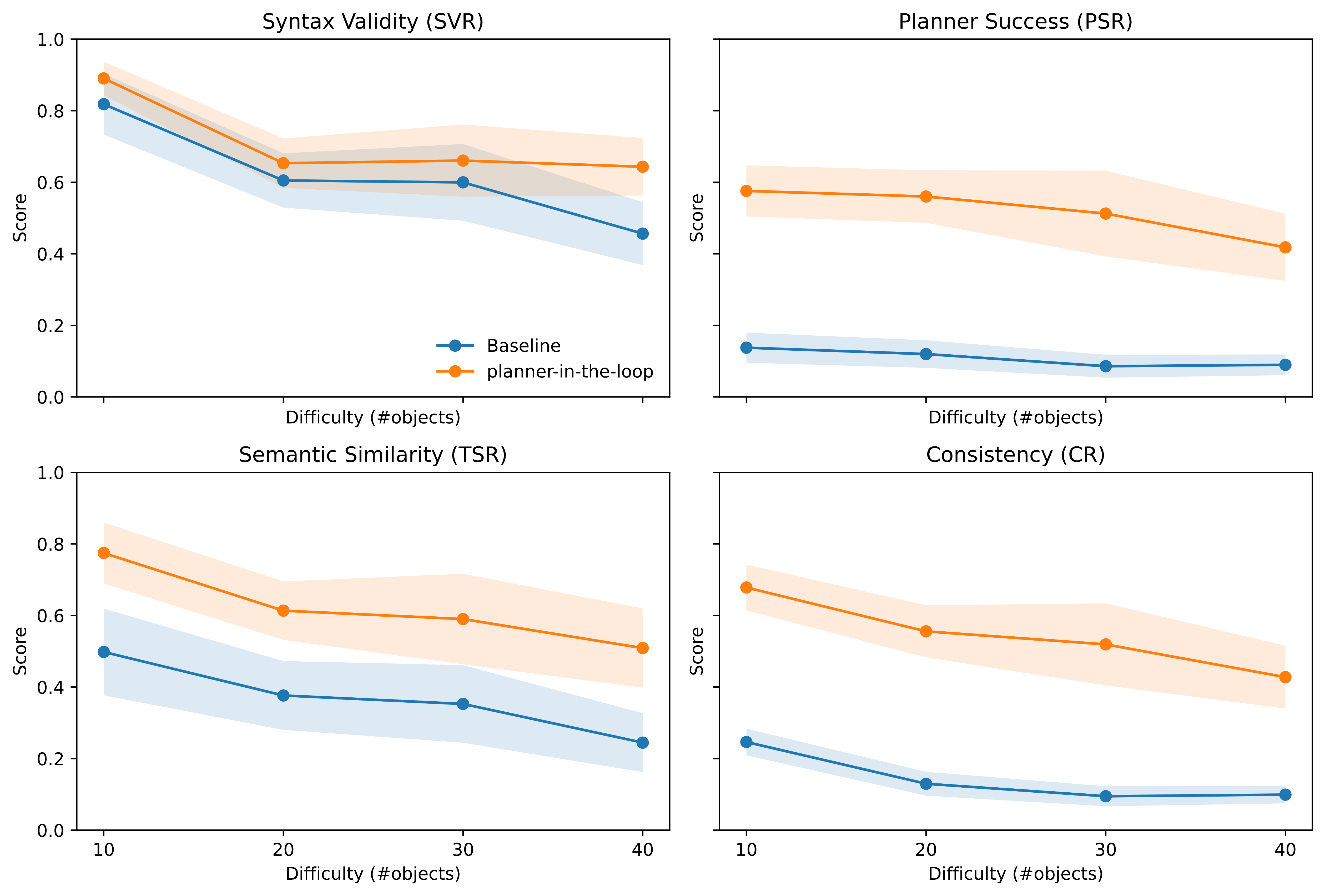}
  \caption{Performance under difficulty scaling.}
  \label{fig:difficulty_scaling}
\end{figure}

To test whether planning reliability is governed by executable constraint satisfaction rather than surface-form fluency, we conduct a difficulty-scaling analysis.As the object scale increases from 10 to 40, the baseline planner success drops from about 13\% to about 9\%, plan-level agreement drops from about 25\% to about 9\%, and specification similarity drops from about 50\% to about 26\%, with the overall score decreasing from about 42\% to about 22\%. In contrast, with planner-in-the-loop feedback, performance remains substantially more stable at the hardest scale: syntax validity is about 64\%, planner success and plan-level agreement are both about 42\%, and the overall score stays near 50\%.

\begin{figure}[htbp]
\centering
\includegraphics[width=\columnwidth]{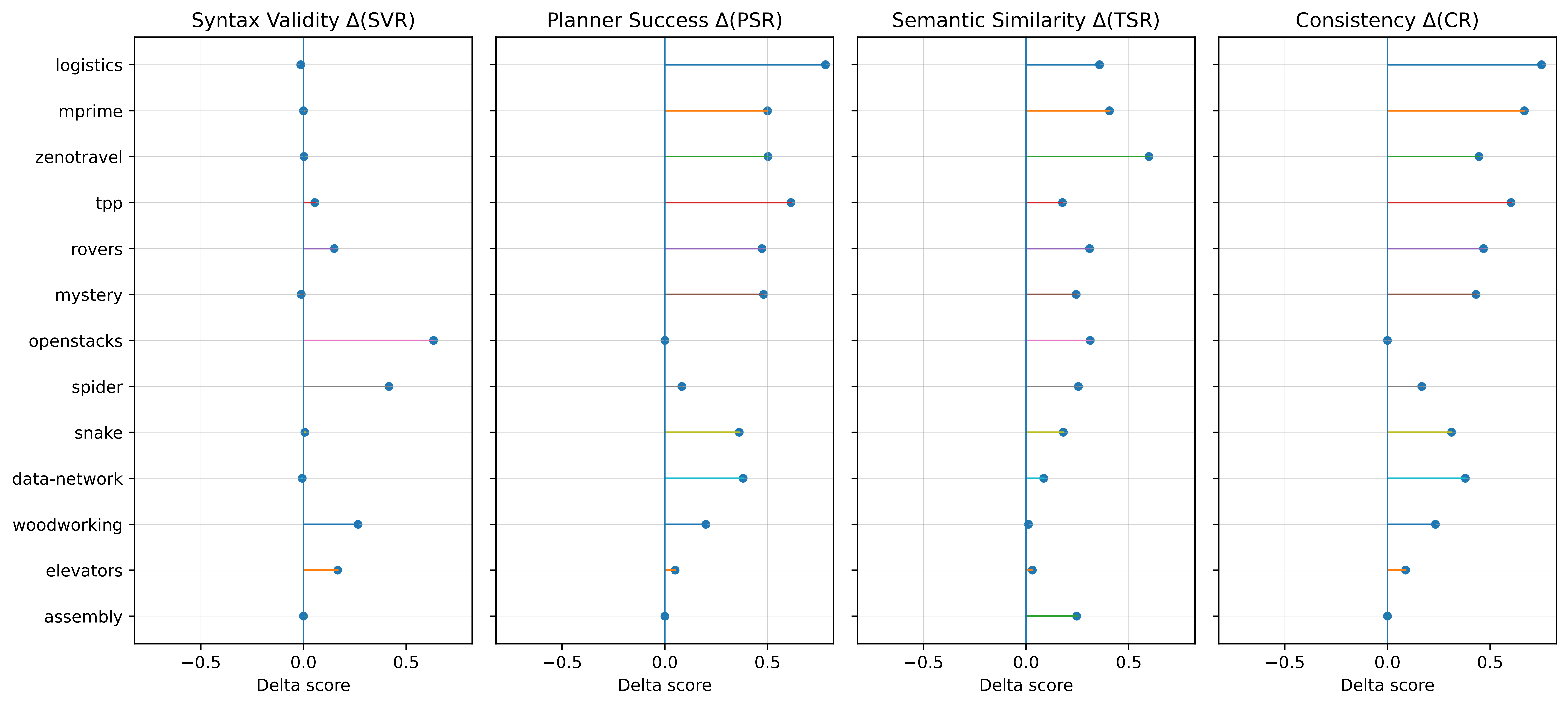}
\caption{Cross-domain generalization results.}
\label{fig:cross_domain_generalization}
\end{figure}

\subsubsection{Cross-Domain Generalization}

We further study cross-domain generalization to assess whether gains reflect domain-specific memorization or a general improvement in executable alignment.In many domains, syntax validity is already relatively high and changes modestly, whereas planner success and plan-level agreement improve substantially, indicating that the main benefit comes from constraint-level alignment. For example, in Logistics, planner success increases from about 13\% to about 78\%, and plan-level agreement increases from about 20\% to about 82\%, consistent with repairing parseable but structurally invalid specifications such as missing connectivity facts or capacity conflicts. In contrast, in resource-intensive domains such as Openstacks, syntax improvements do not yield comparable gains in executability, reinforcing that syntax compliance is not a reliable proxy for executable alignment.

% =========================
\section{Conclusion}

We introduce NL-PDDL-Bench, a multi-domain benchmark for natural-language-to-PDDL specification generation with planner-verified executability and object-count scaling. We present a planner-in-the-loop framework and a planner-grounded optimization recipe, together with a unified evaluation suite for parseability, solvability, similarity, and outcome-aware plan consistency. Across model families and thirteen domains, the method substantially improves planner success and plan-level agreement, with robustness under harder scales and cross-domain variation, supporting more reliable LLM-based planning in safety-sensitive settings.

\section*{Acknowledgment}
This work was supported by the Strategic Priority Research Program of Chinese Academy of Sciences under Grant XDA0480301.

% =========================
% References
% =========================
\bibliographystyle{IEEEtran}
\bibliography{custom}

\end{document}